\documentclass[letterpaper]{article} 
\usepackage{aaai25}  
\usepackage{times}  
\usepackage{helvet}  
\usepackage{amsmath}
\usepackage{multirow,booktabs}
\usepackage{amsfonts}
\usepackage{courier}  
\usepackage[hyphens]{url}  
\usepackage{graphicx} 
\usepackage{natbib}  
\usepackage{caption} 
\usepackage{algorithm}
\usepackage{algorithmic}

\usepackage{newfloat}
\usepackage{listings}
\DeclareCaptionStyle{ruled}{labelfont=normalfont,labelsep=colon,strut=off} 
\floatstyle{ruled}
\newfloat{listing}{tb}{lst}{}
\floatname{listing}{Listing}
\title{AIQViT: Architecture-Informed Post-Training Quantization for \\ Vision Transformers}
\author{
    Runqing Jiang\textsuperscript{\rm 1}, Ye Zhang\textsuperscript{\rm 1}, Longguang Wang\textsuperscript{\rm 2}, Pengpeng Yu\textsuperscript{\rm 1}, Yulan Guo\textsuperscript{\rm 1,}\thanks{Corresponding Author}
}
\affiliations{
    \textsuperscript{\rm 1}Shenzhen Campus, Sun Yat-sen University\\
    \textsuperscript{\rm 2}Aviation University of Air Force\\
    \{jiangrq3,yupp5\}@mail2.sysu.edu.cn, zhangy2658@mail.sysu.edu.cn, wanglongguang15@nudt.edu.cn, guoyulan@sysu.edu.cn

}

\usepackage{bibentry}

\begin{document}

\maketitle

\begin{abstract}
Post-training quantization (PTQ) has emerged as a promising solution for reducing the storage and computational cost of vision transformers (ViTs). Recent advances primarily target at crafting quantizers to deal with peculiar activations characterized by ViTs. However, most existing methods underestimate the information loss incurred by weight quantization, resulting in significant performance deterioration, particularly in low-bit cases. Furthermore, a common practice in quantizing post-Softmax activations of ViTs is to employ logarithmic transformations, which unfortunately prioritize less informative values around zero. This approach introduces additional redundancies, ultimately leading to suboptimal quantization efficacy. To handle these, this paper proposes an innovative PTQ method tailored for ViTs, termed AIQViT (\textbf{A}rchitecture-\textbf{I}nformed Post-training \textbf{Q}uantization for \textbf{ViT}s). First, we design an architecture-informed low-rank compensation mechanism, wherein learnable low-rank weights are introduced to compensate for the degradation caused by weight quantization. Second, we design a dynamic focusing quantizer to accommodate the unbalanced distribution of post-Softmax activations, which dynamically selects the most valuable interval for higher quantization resolution. Extensive experiments on five vision tasks, including image classification, object detection, instance segmentation, point cloud classification, and point cloud part segmentation, demonstrate the superiority of AIQViT over state-of-the-art PTQ methods.
\end{abstract}

%

\section{Introduction}

Thanks to the powerful scalability and superior ability in building long-range dependencies, vision transformers (ViTs) have accomplished cutting-edge performances in various vision tasks, such as image classification \cite{dosovitskiy2020image,touvron2021training}, image object detection \cite{carion2020end,li2023mask}, and point cloud analysis \cite{zhao2021point,guo2021pct}. However, the computational intensity and substantial memory requirements of off-the-shelf ViTs pose significant barriers to their widespread deployment on resource-constrained devices. To this end, a number of model compression techniques have been studied, including network pruning \cite{fang2023depgraph,jiang2022sparse},  knowledge distillation \cite{jiang2023knowledge,zhao2022decoupled}, and model quantization \cite{nagel2022overcoming,xiao2023robustmq}, to reduce the storage and computational cost of complicated models while preserving their performances.

\begin{figure}[t] %
	\centering %
	\includegraphics[width=0.47\textwidth]{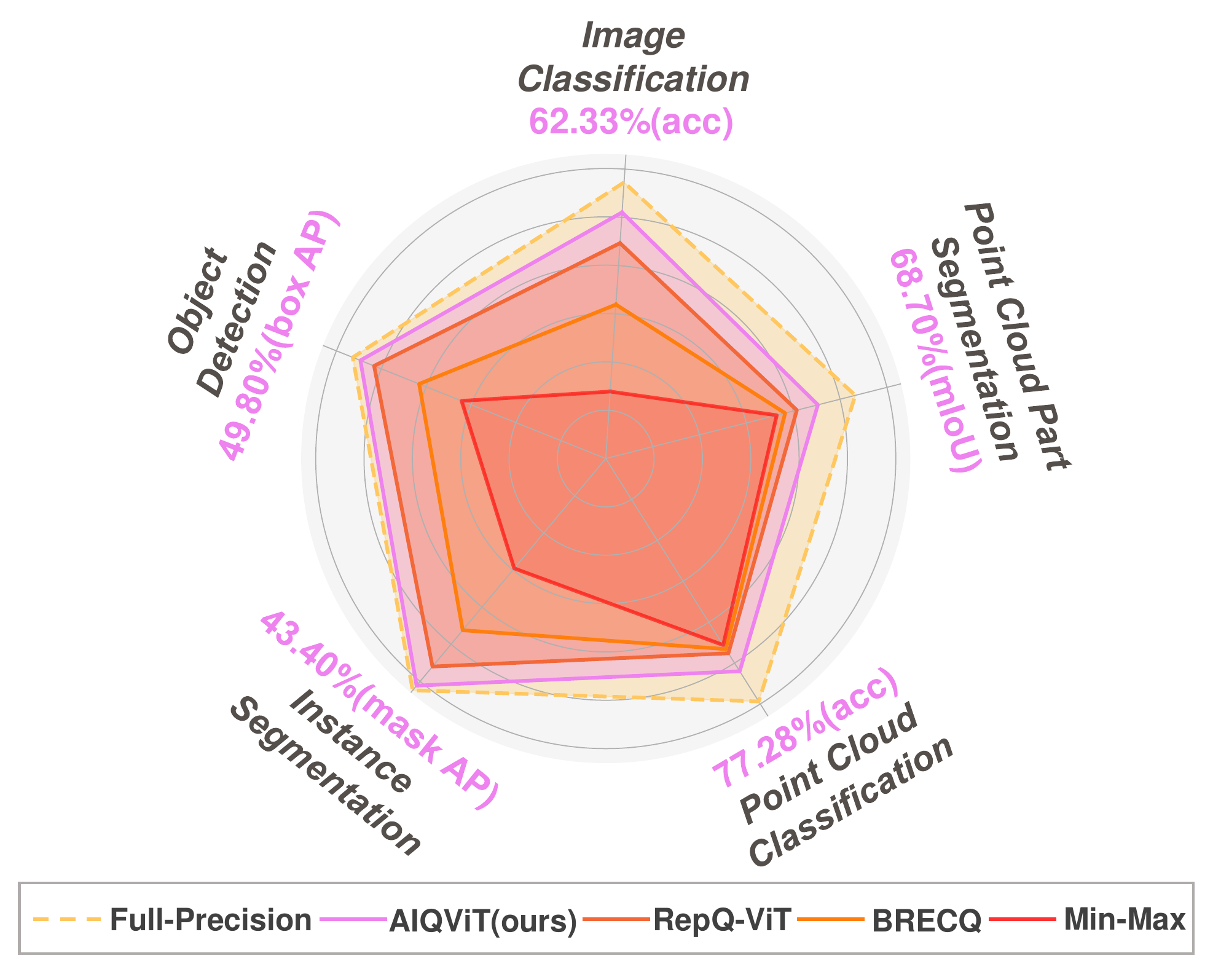} %
	\caption{The performances of different approaches on different tasks (including image classification, object detection, instance segmentation, point cloud classification, and point cloud part segmentation), where the W4/A4 setting is used for quantization. Best viewed in color.} \label{fig:radar} %
	 
\end{figure}

Among the model compression techniques, model quantization has emerged as a widely adopted paradigm, which compresses models by reducing the bit-width of weights and activations. Some quantization-aware training methods \cite{lin2015neural,nagel2022overcoming} require a retraining process to recover accuracy, which is computationally expensive and time-consuming. Moreover, they typically rely on the full training datasets, which may not be feasible due to privacy problems. To handle these, post-training quantization (PTQ) has gained increasing popularity in recent years. Compared with the quantization-aware training methods, PTQ-based methods \cite{li2020brecq,wei2021qdrop} only invocate a small subset of the original training data for calibration, providing a more viable option for model compression.

Early PTQ-based works \cite{li2020brecq,wei2021qdrop} on computer vision tasks primarily target at compressing convolutional neural networks (CNNs), and have obtained remarkable performances. However, it remains challenging to apply PTQ to ViT models due to their unique components (e.g., Softmax) different from those of CNNs, especially in low-bit cases. To this end, some methods \cite{liu2021post,yuan2022ptq4vit,li2023repq} have explored PTQ techniques for ViTs, where specific feature distributions (i.e., post-Softmax and post-LayerNorm) are taken into account. Unfortunately, these works still show limited performances in low-bit cases. This is because the parameter space of full-precision models is not inherently aligned with their quantized counterparts, resulting in substantial quantization errors in weights. Besides, the commonly adopted logarithmic operations tend to preserve precision for zero-around values, which may involve intensive redundant information \cite{meng2022adavit}, disturbing the quantization efficacy.

To address these challenges, this paper proposes AIQViT (\textbf{A}rchitecture-\textbf{I}nformed Post-training  \textbf{Q}uantization for \textbf{ViT}s), a PTQ method tailored for ViTs across different tasks as presented in Figure~\ref{fig:radar}. First, we propose an architecture-informed low-rank compensation mechanism to alleviate the degradation induced by weight quantization. 
Specifically, we introduce learnable weights for each linear layer to remedy the information loss caused by weight quantization. To reduce training cost and prevent overfitting, these weights are designed in low-rank formats, with their ranks determined using network architecture search.
Second, we design a DFQ (\textbf{D}ynamic \textbf{F}ocusing \textbf{Q}uantizer) to handle the post-Softmax activations of ViTs. To be specific, DFQ learns to identify the most valuable interval of post-Softmax activations and subsequently applies standard uniform quantization instead of logarithmic operations within this interval, and thus obtains better quantization efficiency.

In summary, our main contributions are listed as follows:

\begin{itemize}
	\item We propose AIQViT, which is composed of an architecture-informed low-rank compensation mechanism and a dynamic focusing quantizer, to quantize ViTs in a post-training manner, even in low-bit cases. 
 
    \item We develop an architecture-informed low-rank compensation mechanism to compensate for the information loss caused by the weight quantization of ViT models. 
		\item We design a DFQ to deal with the unbalanced distribution of post-Softmax activations without using logarithmic operations, achieving higher quantization efficiency and better performances. 

        
	\item Extensive experiments on five tasks (including image classification, object detection, instance segmentation, point cloud classification, and point cloud part segmentation) with multiple ViT variants validate the superiority of AIQViT against several state-of-the-art PTQ methods.
	 
\end{itemize}

\section{Related Works}

\subsection{Vision Transformers}
Transformer, which arises from natural language processing tasks, has sparked great interest in the field of computer vision \cite{yan2023learning}. ViT \cite{dosovitskiy2020image} pioneeringly introduces a pure transformer architecture on image classification by splitting images into sequences of patches, and attains excellent performance compared to some advanced convolutional neural networks. Later, DeiT \cite{touvron2021training} is proposed, where transformer can be efficiently trained on mid-size image datasets. Inspired by these, some efforts attempt to apply ViTs to other vision tasks, such as object detection \cite{carion2020end,DBLP:conf/nips/FangLWFQWNL21} and point cloud understanding \cite{guo2021pct,zhao2021point}. For instance, DETR \cite{carion2020end} regards object detection as a set prediction task, in which transformers are used for capturing the relationship between objects. After that, YOLOS \cite{DBLP:conf/nips/FangLWFQWNL21} uses an attention-only architecture and sequence-to-sequence strategy for object detection. Additionally, PCT \cite{guo2021pct} is proposed as the first transformer-based backbone for point cloud understanding. PCT takes each point as a token and performs vector attention between points in a local neighbor set, and attains promising performance on point cloud classification and point cloud part segmentation tasks. Generally, the impressive accomplishments of ViTs heavily depend on substantial computation and storage overhead, hindering their applications on devices with limited memory and computation resources. In this paper, we are concerned with the problem of producing quantized ViTs with low-bit weights and activations in a post-training manner. 

\subsection{Model Quantization}
One promising solution to compress complex models and accelerate inference is model quantization. Model quantization decreases the bit-width of weights and activations to alleviate the memory footprint and the computational overhead. Early works \cite{lin2015neural,nagel2022overcoming} commonly adopt the paradigm of quantization-aware training (QAT) to employ retraining on the entire training data for higher accuracy after quantization. Despite their effectiveness, these methods inevitably suffer from privacy problems and massive time overhead, particularly for large-scale ViTs. 

In contrast to QAT, post-training quantization (PTQ), which crafts quantized models without expensive retraining, has attracted increasing attention in model compression. Adaround \cite{DBLP:conf/icml/NagelABLB20} suggests that the naive round-to-nearest is not optimal for quantization, and designs an adaptive rounding strategy to lower the quantization error. After that, BRECQ \cite{li2020brecq} uses basic building blocks in neural networks to perform reconstruction, obtaining impressive image classification accuracy with 4-bit ResNet.
Despite their achievements on CNNs, they show limited performances on ViTs. Thus, many works \cite{DBLP:conf/ijcai/LinZSLZ22, li2023repq} attempt to apply PTQ to ViT models. Due to the uneven activation distributions of post-LayerNorm, RepQ-ViT \cite{li2023repq} uses channel-wise quantization to alleviate the severe inter-channel variations, and then uses reparameterization skills to convert scale factors into layer-wise formats. However, these methods typically rely on log2-based quantizers to deal with the post-Softmax activations, which focus on zero-around values, containing massive redundancy. Besides, these quantizers entail specialized operations to achieve efficiency \cite{lee2017lognet,DBLP:conf/ijcai/LinZSLZ22}. 

\section{Preliminaries}
\subsection{Overview of ViTs}

ViTs are primarily composed of an embedding layer and some stacked transformer blocks, contributing to capturing long-range relationships hidden in different patches. Specifically, an input image $I$ is firstly split into several patches and then fed into a linear layer to obtain its feature representation $X^{(0)} \in \mathbb{R}^{N \times D}$, where $N$ denotes the number of patches and $D$ denotes the embedding dimension. Subsequently, $X^{(0)}$ is sent into a transformer block consisting of a multi-head self-attention (MHSA) followed by a multi-layer perceptron (MLP) module. Mathematically, for the $l$-th transformer block, the above procedure is formulated as: 
\begin{equation}
	\tilde{X}^{(l-1)} = \mathrm{MHSA}^{(l)}(\mathrm{LayerNorm}(X^{(l-1)})) + X^{(l-1)} 
\end{equation}
\begin{equation}
	X^{(l)} = \mathrm{MLP}^{(l)}(\mathrm{LayerNorm}(\tilde{X}^{(l-1)})) + \tilde{X}^{(l-1)}.
\end{equation}
The MHSA module attends to all the patches by:
\begin{equation}
	\mathrm{Attn}_i = \mathrm{Softmax} (\frac{Q_iK_i^T}{\sqrt{D_h}})V_i
\end{equation}
\begin{equation}
	{\rm MHSA}(X) = \mathrm{concat}(\mathrm{Attn}_1, \mathrm{Attn}_2, ..., \mathrm{Attn}_H)W_o
\end{equation}
where $Q_i = XW^q_i$, $K_i = XW^k_i$, $V_i = XW^v_i$ respectively represent the query, key and value for the $i$-th head. $D_h$ denotes the feature dimension for each head. $H$ is the number of attention heads. For readability, we omit the bias term. Then, the MLP encodes the patch features individually, which is defined as:
\begin{equation}
	\mathrm{MLP}(X) = \mathrm{GELU}(XW_1)W_2
\end{equation} 
where $W_1$ and $W_2$ denote the weights in linear layers.

\subsection{Quantizers}
\textbf{Uniform Quantizer}.
It is widely used due to its broad compatibility across hardware platforms and is defined as:
\begin{equation}
	x_q = \mathrm{Quant-U}(x) = \mathrm{clamp}(\big \lfloor \frac{x}{s} \big \rceil + z, 0, 2^k-1)
\end{equation}
where the $x$ is a floating-point value and $x_q$ is the quantized value. The $s$, $z$, and $k$ represent the quantization scale, zero point, and bit-width, respectively. In line with previous works \cite{li2023repq, zhong2023s}, we employ channel-wise quantization for weights and layer-wise quantization for activations during inference. 

\noindent \textbf{Log2-Based Quantizer}.
It is specifically designed to address the challenge of non-negative long-tail activation quantization, which can be mathematically formulated as:
\begin{equation}
	x_q = \mathrm{Quant-LU}(x) = \mathrm{clamp}(\big \lfloor-\mathrm{log}_2(\frac{x}{s}) \big \rceil, 0, 2^k - 1).
\end{equation}

\subsection{Low-Rank Adaptation (LoRA)}
Low-Rank Adaptation (LoRA) \cite{DBLP:conf/iclr/HuSWALWWC22} is broadly studied in the realm of large language models to achieve parameter-efficient fine-tuning. Mathematically, given a pre-trained weight matrix $W_0 \in \mathbb{R}^{m \times d}$, the process of LoRA is:
\begin{equation}\label{eq:lora}  
	h = xW_0 + xBA
\end{equation}
where $h\in\mathbb{R}^{d}$ is the hidden state corresponding to input $x\in\mathbb{R}^m$, $B \in \mathbb{R}^{d \times r}$, $A \in \mathbb{R}^{r \times m}$, and the rank $r << \min(d, m)$. During training, only the $A$ and $B$ are updated by gradient descent while remaining $W_0$ frozen. After training, the $BA$ is seamlessly merged into $W_0$ by assigning $W_0 = W_0 + BA$. 

\section{Method}
\subsection{Architecture-Informed Low-Rank Compensation}

\begin{figure}[t]
	\centering 
	\includegraphics[width=0.47\textwidth]{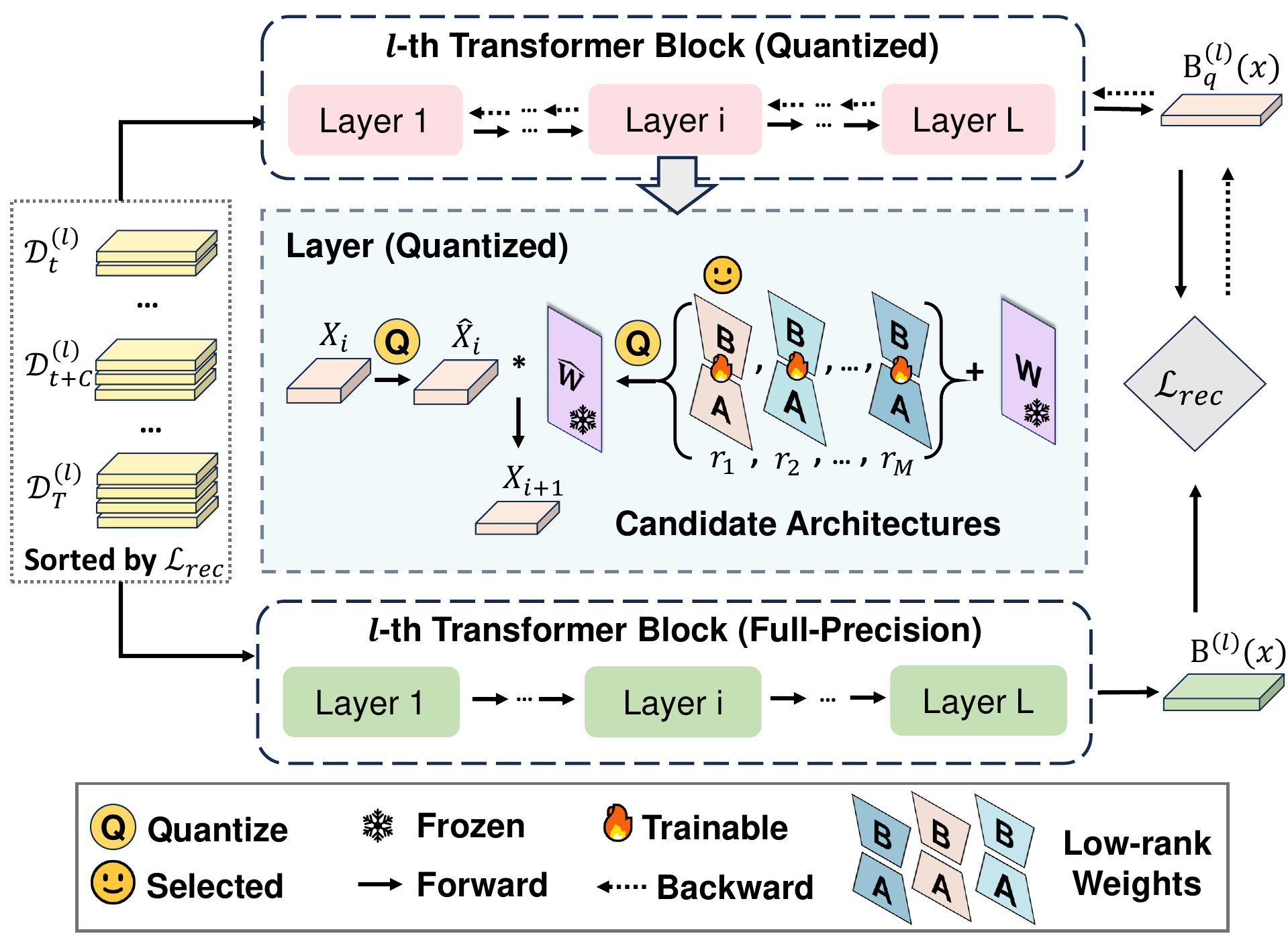} 
	\caption{Overview of architecture-informed low-rank compensation. First, we employ differential architecture search to identify the optimal rank $r$ from a candidate architecture set. Subsequently, we freeze the original weights and optimize the selected low-rank weights by minimizing the reconstruction loss between the full-precision block and the quantized block. During this process, the training set is incrementally expanded in a curriculum learning (CL) manner.} \label{fig:arloc} 
\end{figure}

Unlike CNNs, ViTs are composed of a large number of fully connected (FC) layers, necessitating a significant amount of computing and storage resources. However, due to the more intricate architectures of ViTs, directly applying weight quantization for these layers may incur significant information loss, leading to inferior accuracy. To address this, some learnable weights are introduced for FC layers to compensate for information loss. These learnable weights are formulated in low-rank formats, effectively reducing optimization costs while preventing overfitting that may arise from limited data.
The process is shown in Figure~\ref{fig:arloc}.

\noindent \textbf{The Process of Low-Rank Compensation.}
Inspired by LoRA \cite{DBLP:conf/iclr/HuSWALWWC22}, we introduce learnable low-rank weights into quantization and reformulate Eq.~(\ref{eq:lora}):
\begin{equation}
	h = \text{Quant-U}(x)\text{Quant-U}(W_0 + BA)
\end{equation}
where we hold $W_0$ fixed and update $A$ and $B$ to minimize the following reconstruction loss:
	\begin{equation}\label{eq:block}
		\mathcal{L}_{rec}(\mathcal{D}^{(l)}_t, \Theta^{(l)}) = \mathbb{E}_{x \in \mathcal{D}^{(l)}_t} \left( \left\| \text{B}_q^{(l)}(x) \newline  - \text{B}^{(l)}(x) \right\|_{\text{F}} \right)
	\end{equation}
where $\mathcal{D}^{(l)}_t$ denotes the input dataset for the $l$-th block at $t$-th iteration, $ \Theta^{(l)}$ denotes all the learnable low-rank weights for the $l$-th block, $\left\| \cdot \right\|_{\text{F}}$ represents the Frobenius norm, $\text{B}^{(l)}(x)$ and $\text{B}_q^{(l)}(x)$ represent the outputs of the $l$-th full-precision block and quantized block, respectively. Thanks to the learnable low-rank weights, the quantized model is encouraged to learn a parameter space which is compatible for quantization, and the reconstruction error induced by weight quantization is significantly alleviated without incurring massive optimization overhead. During reconstruction, channel-wise quantization is used for post-LayerNorm activations. After that, we use scale reparameterization \cite{li2023repq} to convert them into layer-wise quantization.

\noindent \textbf{The Choice of Rank $r$.}
Since the choice of $r$ has a profound impact on the performance of quantized models, we are motivated to automate the search for $r$. Specifically, we model the search of $r$ as a network architecture search problem, and thus use efficient differentiable architecture search to handle this issue. Given a proposal set $\mathcal{S} = \{r_1, r_2, \cdots, r_M\}$ for the rank $r$, the forward pass of a fully-connected layer is:
\begin{equation}
	y = \hat{x}W_0 + \sum_{r_j \in \mathcal{S}} \phi(\overline{\alpha}_{r_j}\hat{x}B_{r_j}A_{r_j}),
\end{equation},
\begin{equation}
\overline{\alpha}_{r_j} = \frac{\exp({\alpha}_{r_j})}{\sum_{i=1}^{M}\exp({\alpha}_{r_i})} 
\end{equation}
where $\hat{x}=\text{Quant-U}(x)$ represents the quantized value of input $x$, and $B_{r_j} \in \mathbb{R}^{d \times r_j}$, $A_{r_j} \in \mathbb{R}^{r_j \times m}$ are candidate weights with respect to $r_j$. $\alpha = \{\alpha_{r_i}\}_{i=1}^{M}$ are a set of learnable parameters to control the importance of each architecture, and the $\phi(\cdot)$ denotes drop-path operation \cite{larsson2016fractalnet}. Meanwhile, we divide the calibration set $\mathcal{D}$ into a training set $\mathcal{D}_{train}$ and a validation set $\mathcal{D}_{val}$, and define the optimization objective:
\begin{equation}
	\min\limits_{\alpha} \quad \mathcal{L}_{rec}(\mathcal{D}^{(l)}_{val}, \Theta^{(l)}_{*})
\end{equation}
\begin{equation}
	\text{s.t. } \Theta^{(l)}_{*} = {\arg\min}_{\Theta^{(l)}} \mathcal{L}_{rec}(\mathcal{D}_{train}^{(l)}, \Theta^{(l)}).
\end{equation}
The above bilevel optimization problem can be efficiently solved by approximate architecture gradient used in \cite{liu2018darts}. After that, for each linear layer, the optimal rank is determined by $r^* = \arg\max_{r\in\mathcal{S}}{\alpha}_r$.

Note that, albeit our AIQViT and QLLM \cite{liu2023qllm} use LoRA to compensate for quantization error, the differences between AIQViT and QLLM are significant. First, QLLM aims to quantize large language models while our AIQViT is designed for ViTs. Second, QLLM sets the value of $r$ empirically. However, it is found that the rank $r$ is critical for the final performance. Thus, we determine $r$ by using network architecture search instead of manual settings.

\subsection{Dynamic Focusing Quantizer}
\begin{figure}[!t]
	\centering 
	\includegraphics[width=0.47\textwidth]{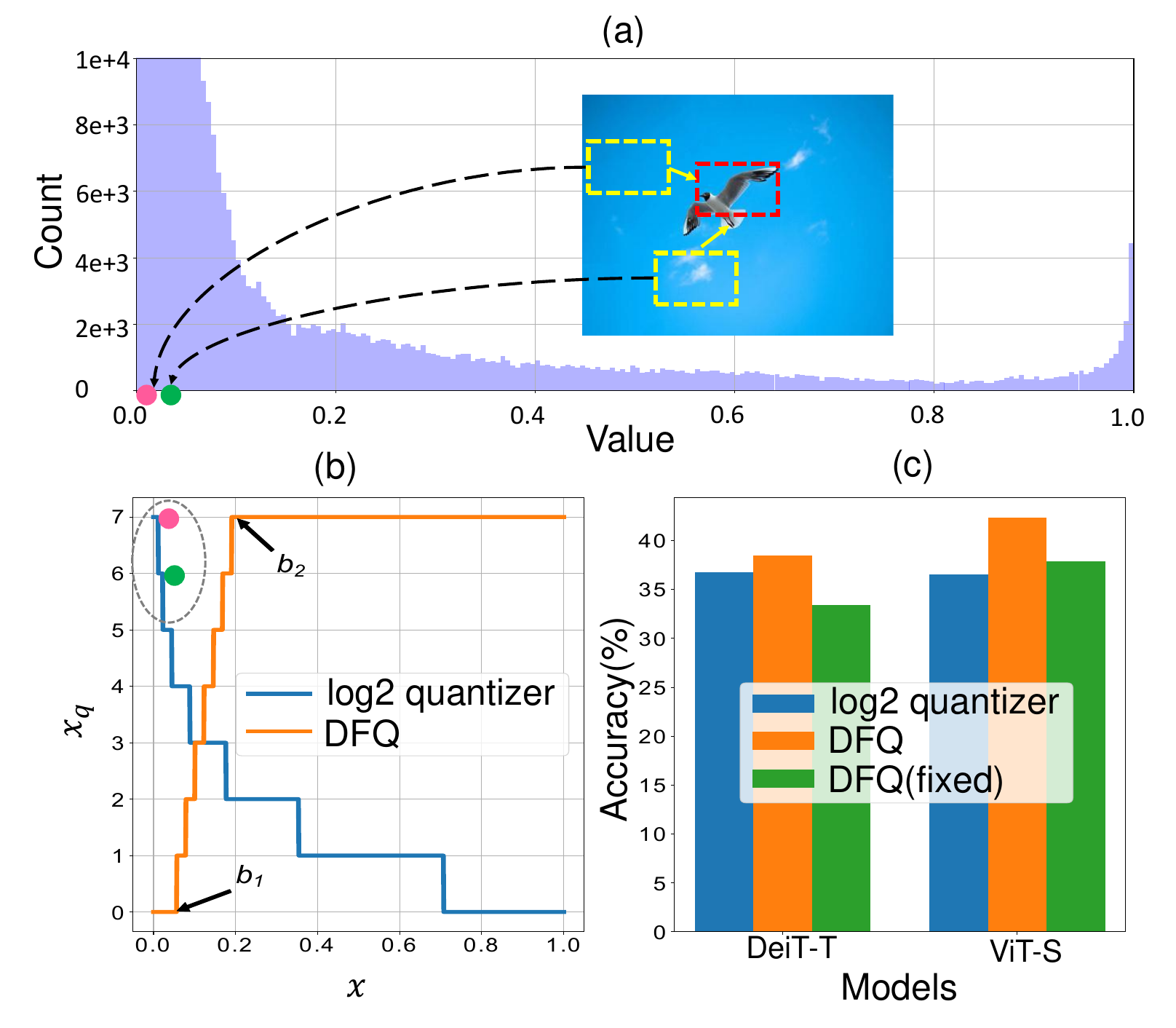} 
	\caption{(a) Histogram of the first MHSA module’s post-Softmax activations in DeiT-T. (b) log2 quantizer (in blue) and DFQ (in orange). (c) Results on ImageNet with W3/A3 quantization. ``DFQ(fixed)'' means all the layers use the same interval. Best viewed in color.} \label{fig:dfq} 
\end{figure}

As indicated in Figure~\ref{fig:dfq}(a), the post-Softmax activations exhibit unbalanced distribution, which has become one of the central challenges in quantizing ViTs. To handle this, several methods \cite{li2023repq,DBLP:conf/ijcai/LinZSLZ22} leverage the non-linearity of logarithmic operation and adopt log2-based quantizers to quantize the post-Softmax activations. 
Here, we visualize the execution process of a log2 quantizer in Figure~\ref{fig:dfq}(b). It can be observed that inputs with smaller values generally correspond to higher quantization resolution and vice versa. Such a schema indicates that the most valuable activations are located within a partial interval instead of the whole one. However, the log2 quantizer tends to preserve values around 0, which may contain a plethora of redundant information, disturbing the efficacy of quantization. Besides, the log2 quantizer keeps the interval fixed for each layer, which may not be the optimal solution. Based on the above insights, we design a DFQ (\textbf{D}ynamic \textbf{F}ocusing \textbf{Q}uantizer) to select the most valuable interval for each Softmax layer dynamically. To be specific, let $\left[b_1, b_2 \right]$ be the selected interval where $b_1$ and $b_2$ are learnable parameters, which is:
\begin{equation}
		\begin{split}
			x_{q}= \left \{ 
			\begin{array}{ll}
				0, & x < b_1\\
				\text{Quant-U}(x), & {b_1} \leq x \leq {b}_2 \\
				2^k-1, & x > {b}_2
			\end{array}
			\right.
		\end{split}
\end{equation}
where $x\in\left[0, 1\right]$ is the activation after Softmax and $k$ is the bit-width. By dynamically selecting ${b}_1$ and ${b}_2$, DFQ will focus on the most valuable interval and prioritize more bits accordingly. Then, we use a uniform quantizer for values in  $\left[ {b}_1, {b}_2 \right]$, and directly convert the values in $[0, {b}_1)$ and $({b}_2, 1]$ to 0 and $2^k-1$, respectively. In this way, post-Softmax activations are quantized without logarithmic operations, and we find DFQ can achieve comparable results with log2-based quantizers, as illustrated in Figure~\ref{fig:dfq}(c). 

\subsection{Optimization Strategy}
To minimize Eq.~(\ref{eq:block}), we deviate from the standard training pipeline employed in BRECQ \cite{li2020brecq}. That is because we observe that arranging samples in a meaningful order yields better performances, especially in ultra-low bit case (i.e., 3-bit). Inspired by curriculum learning, we optimize Eq.~(\ref{eq:block}) by firstly feeding easier samples, and then gradually introducing the harder ones. To be specific, reconstruction loss is selected to gauge the hardness of each sample, and the training data for the $t$-th iteration is:
\begin{equation}
	\mathcal{D}^{(l)}_t = {\arg\min}_{\hat{\mathcal{D}}:|\hat{\mathcal{D}}| \geq \lambda(t) \cdot |\mathcal{D}|} \mathcal{L}_{rec}(\hat{\mathcal{D}}, \Theta^{(l)})
\end{equation}
where $\mathcal{D}$ denotes the calibration data, the $|\mathcal{D}|$ and $|\hat{\mathcal{D}}|$ represent the size of $\mathcal{D}$ and $\hat{\mathcal{D}}$. $\lambda(t)$ is used to schedule the proportion of training samples at the $t$-th iteration, which is defined by a linear function \cite{DBLP:journals/pami/WangCZ22}:
\begin{equation}
	\lambda(t) = \min\big ( 1, \lambda_0 + \frac{1-\lambda_0}{T}\cdot t\big)
\end{equation} 
where $\lambda_0$ is the initial proportion of training samples which is set to 0.5 in our method, and $T$ is the total number of iterations. Thus, the quantized model tends to learn high-confidence regions in the early training stage, mitigating the negative impact of outliers and preparing a more favorable parameter space for stable optimization.

\section{Experiment}
\subsection{Experimental Setups}

\noindent \textbf{Datasets and Competing Methods}. 
We adopt ImageNet \cite{krizhevsky2012imagenet}, ModelNet40 \cite{wu20153d}, and ShapeNetPart \cite{yi2016scalable} for image classification, point cloud classification, and point cloud part segmentation respectively. The COCO \cite{lin2014microsoft} dataset is used to evaluate object detection and instance segmentation tasks. We use BRECQ \cite{li2020brecq}, QDrop \cite{wei2021qdrop}, PTQ4ViT \cite{yuan2022ptq4vit}, RepQ-ViT \cite{li2023repq}, APQ-ViT \cite{ding2022towards}, Min-Max \cite{DBLP:conf/icml/NagelABLB20} as competing methods.

\noindent \textbf{Implementation Details.}
For image classification, we adopt the experimental settings from \cite{zhong2023s} and randomly select a calibration set of 1,024 samples from the ImageNet dataset. We evaluate our method on a variety of models, including ViT \cite{dosovitskiy2020image}, DeiT \cite{touvron2021training}, and Swin Transformer \cite{liu2021swin}, with batch size set to 24. For object detection and instance segmentation, we employ Mask-RCNN \cite{he2017mask} and Cascade Mask-RCNN \cite{cai2018cascade} on the COCO dataset, using a calibration set of 256 samples and a batch size of 1. For point cloud classification and part segmentation, we evaluate our method on Point Transformer \cite{zhao2021point} with a batch size of 32 and a calibration set of 512 samples. The iteration numbers are set to 2,000 and 6,000 for network architecture search and calibration, respectively. We empirically set $\mathcal{S}=\{10, 20, 50, 100, 150\}$ for all model variants across all vision tasks.

\subsection{Image Classification on ImageNet}
Table~\ref{tab:imagenet} shows the test accuracy results obtained by various competing methods on ImageNet with different models and bit-widths. Notably, in the case of W3/A3 quantization, most competing methods exhibit unfavorable test accuracy on all model variants. For instance, RepQ-ViT achieves only 0.44\% and 0.17\% top-1 accuracy in ViT-S and ViT-B quantization, respectively, which is far from achieving practical usability. In contrast, BRECQ and QDrop perform slightly better than RepQ-ViT by leveraging block-wise reconstruction, which enables more accurate quantization parameters to be obtained. In the W4/A4 quantization setting, APQ-ViT achieves accuracy of 47.94\% and 43.55\% in DeiT-T and DeiT-S quantization, respectively, while RepQ-ViT improves test accuracy by over 9\% owing to the application of scale reparameterization skills. Furthermore, in the W6/A6 quantization setting, all competing methods attain satisfied performance, with minimal performance gaps across different methods. Nonetheless, our AIQViT achieves the highest top-1 accuracy in most scenarios. For example, compared to the Full-Precision DeiT-B, our AIQViT achieves 81.40\% test accuracy in the case of W6/A6 quantization, which corresponds to a mere 0.4\% accuracy loss.

\begin{table*}[t]
	\centering
	\renewcommand{\arraystretch}{1}
	\begin{tabular}{cccccccccc}
	\toprule
		{\textbf{Method}}   &\textbf{REC} &\textbf{Bits.(W/A)}  & \textbf{DeiT-T} &   \textbf{DeiT-S} &   \textbf{DeiT-B} & \textbf{ViT-S} & \textbf{ViT-B} & \textbf{Swin-S} & \textbf{Swin-B} \\
	\midrule 
		Full-Precision         &-      & {32/32}      & 72.21                      & 79.85                      & 81.80                      & 81.39                     & 84.54                     & 83.23                      & 85.27                      \\
	\midrule 
		RepQ-ViT  &$\times$ &3/3                             &1.15                            &4.85                            &7.23                            &0.44                           &0.17                           &1.22                            &4.87                            \\
        QDrop  &$\checkmark$  &3/3                             & 15.68                           & 16.89                           &  22.38                          & 4.32                          & 5.99                           & 58.96                           & 52.19                           \\
        
         BRECQ &$\checkmark$  &3/3                             & 14.12                           & 8.26                           & 12.87                           &  1.86                         & 0.14                          & 7.32                           & 1.21                           \\
		AIQViT (ours) &$\checkmark$  &3/3                             &\textbf{38.51}                            &\textbf{55.36}                            &\textbf{66.15}                            &\textbf{41.32}                           &\textbf{43.68}                           &\textbf{71.42}                            &\textbf{63.01}                            \\
	\midrule 
		
		PTQ4ViT &$\checkmark$  &4/4                              &36.96                            &34.08                            &64.39                            & 42.57              &30.69                           & 76.09                  &74.02                            \\
		
		RepQ-ViT  &$\times$  &4/4    &57.43                            &69.03                            &75.61                            & 65.05                           &68.48                           &79.45                            &78.32 \\

            APQ-ViT &$\checkmark$ &4/4 &47.94 &43.55 &67.48 &47.95 &41.41 &77.15 &76.48 \\

            QDrop &$\checkmark$   &4/4                            &31.65                            & 35.79                           &65.47                           &17.77                           & 21.72                           & 78.92 &80.49                           \\
        BRECQ &$\checkmark$ &4/4               & 51.60                           &54.31                            &62.96                            & 63.90                 &61.54                           &76.63                    &74.15                            \\
		AIQViT (ours) &$\checkmark$ &4/4                              &\textbf{62.33}                            &\textbf{72.75}                            &\textbf{79.19}         &\textbf{70.63}                           &\textbf{74.15}                           &\textbf{80.93}                           &\textbf{81.22} \\
		\midrule 
	
	PTQ4ViT  &$\checkmark$ &6/6                              &69.68                            &76.28         &80.25                            &78.63                           &81.65                           &82.38                            & 84.01                           \\
	
	RepQ-ViT &$\times$ &6/6                              &70.76                            &78.90                            &81.27                            &\textbf{80.43}                           &83.62                 &82.79                            &\textbf{84.57} \\

    APQ-ViT &$\checkmark$ &6/6 &70.49 &77.76 &80.42 &79.10 &82.21 &82.67 &84.18 \\ 

  QDrop  &$\checkmark$  &6/6                        &70.61                            &78.91                            &80.19                            &73.01                           &80.86                           &81.92                            &83.85                            \\
  
        BRECQ &$\checkmark$  &6/6            &69.70                            &73.09                            &75.67                            &79.61                    &77.19                           &77.66                    &78.03                            \\
	AIQViT (ours)                               & $\checkmark$          &6/6                 &\textbf{70.78}                            &\textbf{78.98}                            &\textbf{81.40}                           &{80.21}                           &\textbf{83.68}                            &\textbf{82.81} &84.39 \\
	\bottomrule                      
	\end{tabular}
 \caption{Quantization results on ImageNet dataset. The top-1 accuracy (\%) is reported as the evaluation metric. ``Bits.(W/A)'' represents the bit-width for weight and activation. ``REC'' means ``Reconstruction''. } \label{tab:imagenet}
\end{table*}

\subsection{Object Detection and Instance Segmentation on COCO}
\begin{table*}[!tb]
	\centering
	
	\begin{tabular}{ccccccccccc}
		\toprule
		\multirow{3}{*}{\textbf{Method}} &\multirow{3}{*}{\textbf{REC}} & \multirow{3}{*}{\textbf{Bits.(W/A)}} & \multicolumn{4}{c}{\textbf{Mask R-CNN}}     &\multicolumn{4}{c}{\textbf{Cascade Mask R-CNN}}                         \\
		&      &                     & \multicolumn{2}{c}{\textbf{w.Swin-T}} & \multicolumn{2}{c}{\textbf{w.Swin-S}} &  \multicolumn{2}{c}{\textbf{w.Swin-T}} & \multicolumn{2}{c}{\textbf{w.Swin-S}}\\
		&              &             & $\text{AP}^{\text{box}}$            & $\text{AP}^{\text{mask}}$           & $\text{AP}^{\text{box}}$             & $\text{AP}^{\text{mask}}$   & $\text{AP}^{\text{box}}$            & $\text{AP}^{\text{mask}}$           & $\text{AP}^{\text{box}}$             & $\text{AP}^{\text{mask}}$           \\
		\hline
		Full-Precision       &-   & 32/32                         & 46.0          & 41.6         & 48.5          & 43.3    &50.4 &43.7 &51.9 &45.0     \\
		\hline
		PTQ4ViT       &$\checkmark$        & 4/4  & 6.9     & 7.0          & 26.7          & 26.6 &14.7 &13.5       &0.5 &0.5 \\
		APQ-ViT       &$\checkmark$          & 4/4                       & 23.7          & 22.6         & \textbf{44.7}          & 40.1     &27.2 &24.4 &47.7 &41.1    \\
		RepQ-ViT      &$\times$          & 4/4                       & 36.1          & 36.0         & 44.2          & 40.2  &47.0 &{41.4}      &49.3 &43.1 \\
		BRECQ         &$\checkmark$      & 4/4                       & 33.7          & 33.1         & 39.5          & 37.7         
 & 42.1 &36.3 &43.2 &37.6\\
		AIQViT (ours)     &$\checkmark$       & 4/4                       & \textbf{38.2}          & \textbf{36.7}         & 44.1          & \textbf{40.4}      &\textbf{47.1} &\textbf{41.4}   &\textbf{49.8} &\textbf{43.4}\\
		\hline
		PTQ4ViT       &$\checkmark$          & 6/6                       & 5.8           & 6.8          & 6.5          & 6.6  &14.7 &13.6 &	12.5 &10.8       \\
		APQ-ViT       &$\checkmark$          & 6/6                       & \textbf{45.4}          & {41.2}         & \textbf{47.9}          & 42.9 &48.6 &42.5 &50.5 &43.9         \\
		RepQ-ViT     &$\times$            & 6/6                       & 45.1          & {41.2}         & 47.8          & {43.0}    &50.0 &43.5      &51.4 &{44.6}\\
		BRECQ         &$\checkmark$       & 6/6                       & 36.1          & 36.0         & 44.2          & 40.2        &46.9 &39.9 &48.3 &41.9 \\
		AIQViT (ours)   &$\checkmark$         & 6/6                       &45.3          & \textbf{41.2}         & 47.5          & \textbf{43.0}  &\textbf{50.2} &\textbf{43.6}    &\textbf{51.5} &\textbf{44.6}  \\
		\bottomrule
	\end{tabular}
 \caption{Quantization results on COCO dataset. ``$\text{AP}^{\text{box}}$'' means the box average precision for object detection, and ``$\text{AP}^{\text{mask}}$'' means the mask average precision for instance segmentation.} \label{tab:coco}
\end{table*}
In line with RepQ-ViT \cite{li2023repq}, we evaluate the performance of object detection and instance segmentation tasks on the COCO dataset, with the quantization results shown in Table~\ref{tab:coco}.  Notably, PTQ4ViT exhibits the worst performance across different quantization settings, which can be attributed to the loss of generality of its exquisite twin-scale mechanism when applied to more complex architectures. Furthermore, our experiments reveal that APQ-ViT achieves desired results when employing Swin-S as the backbone, but suffers from significant performance degradation when using Swin-T as the backbone, indicating a lack of robustness with respect to the choice of backbone. In contrast, our AIQViT demonstrates superior performance when adopting W4/A4 quantization on Mask R-CNN with Swin-S as the backbone, outperforming BRECQ by 4.6 box AP and 2.7 mask AP. Notably, for the W6/A6 quantization of Cascade Mask R-CNN with Swin-T as the backbone, our AIQViT obtains a box AP of 50.2 and a mask AP of 43.6, which is very close to its full-precision counterpart, with a mere 0.2 box AP and 0.1 mask AP gap.

\subsection{Point Cloud Classification on ModelNet40}
\begin{table}[!h]
	\centering
 
	\begin{tabular}{ccc|cc}
		\toprule
		\multirow{1}{*}{\textbf{Method}} &\textbf{REC} & \multirow{1}{*}{\textbf{Bits.(W/A)}}                  & \textbf{mAcc}           & \textbf{OA}     \\
		\midrule
		Full-Precision        &-  & 32/32                         & 89.67         & 92.38          \\
		\midrule
		Min-Max          &$\times$      & 4/4                       &70.60               &73.99                            \\
   RepQ-ViT           &$\times$    & 4/4                                     &72.61                  &77.98        \\
   
		BRECQ            &$\checkmark$       & 4/4                       &71.58               &76.68                  \\
		AIQViT(ours)    &$\checkmark$        & 4/4                       &\textbf{77.28}               &\textbf{82.66}                       \\
		\midrule
		Min-Max        &$\times$        & 6/6                       &83.24               &86.86                          \\
   RepQ-ViT           &$\times$    & 6/6                                     & 85.99                 & 89.17       \\
		BRECQ               &$\checkmark$    & 6/6                                      &85.27                &88.99                \\
		AIQViT(ours)      &$\checkmark$      & 6/6                                      &\textbf{87.31}                &\textbf{90.60}                  \\
		\bottomrule
  	
	\end{tabular}
 \caption{Results on ModelNet40. ``mACC'' and ``OA'' are the mean of class-wise and overall accuracy (\%), respectively. } \label{tab:modelnet} 
\end{table}
We also demonstrate the effectiveness of AIQViT on the 3D point cloud classification task. Specifically, we select Point Transformer \cite{zhao2021point} as the backbone and validate our method on ModelNet40. The experimental results are shown in Table~\ref{tab:modelnet}. As can be observed, the Min-Max obtains the worst performances on both W4/A4 and W6/A6 quantization, which is mainly due to the inaccurate quantization parameters estimated by min-max criteria. Besides, our AIQViT outperforms BRECQ by 5\% in W4/A4 quantization, indicating that AIQViT can be well applied to 3D point cloud classification tasks.

\subsection{Point Cloud Part Segmentation on ShapeNetPart}
For the 3D point cloud part segmentation tasks, the superiority of AIQViT is validated on the ShapeNetPart dataset with mean intersection
over union (mIoU) used as the evaluation metric. Experimental results are provided in Table~\ref{tab:shapenetpart}. As presented in Table~\ref{tab:shapenetpart}, it can be easily observed that AIQViT consistently outperforms other competing methods on W4/A4 and W6/A6 quantization. Specifically, for the W4/A4 case, AIQViT outperforms BRECQ by a large margin. In the case of W6/A6 quantization, AIQViT achieves 76.99 c.mIoU and 81.76 i.mIoU, which is only 1.58 c.mIoU and 1.50 i.mIoU lower than the Full-Precision model. 
   
\begin{table}[]
	\centering
	
	\setlength{\tabcolsep}{1.5mm}{
	\begin{tabular}{ccc|cc}
		\toprule
		\multirow{1}{*}{\textbf{Method}} &\textbf{REC} & \multirow{1}{*}{\textbf{Bits.(W/A)}}      & \textbf{c.mIoU} & \textbf{i.mIoU} \\
		\midrule
		Full-Precision          & -          &32/32    & 78.57    & 83.26        \\
		\midrule
		Min-Max           &$\times$    & 4/4                                     &61.49                  &66.05        \\
        RepQ-ViT           &$\times$    & 4/4                                     & 65.78                 &68.43        \\
		BRECQ                 &$\checkmark$  & 4/4                       &63.26               &66.86                                  \\
		AIQViT(ours)       &$\checkmark$     & 4/4                  &\textbf{68.27}          &\textbf{73.75}              \\
		\midrule
		Min-Max           &$\times$     & 6/6                       &72.88               &76.98                              \\
   RepQ-ViT           &$\times$    & 6/6                                     & 74.58                 & 79.93       \\
		BRECQ            &$\checkmark$       & 6/6                       &73.14               &79.79                            \\
		AIQViT(ours)    &$\checkmark$        & 6/6                  &\textbf{76.99}          &\textbf{81.76}             \\
		\bottomrule
	\end{tabular}
}
\caption{Results on ShapeNetPart. ``c.mIoU'' and ``i.mIoU'' respectively denote the category mIoU and instance mIoU. }\label{tab:shapenetpart}
\end{table}

\subsection{Ablation Studies}
\begin{table}[!h]
	\centering
	
	\begin{tabular}{ccc|ccc}
		\toprule
		\multicolumn{3}{c}{\textbf{Method}} & \multicolumn{3}{c}{\textbf{Bits.(W/A)}} \\
		\midrule
		AILoC     & DFQ    & CL    & 3/3      & 4/4     & 6/6     \\
		\midrule
		$\times$ & $\times$        &$\times$        &14.70           &46.58         &61.24         \\
		$\checkmark$ &$\times$        &$\times$       &30.01          & 57.38        &68.85         \\
		$\times$       &$\checkmark$       & $\times$   &21.43       &50.28         &63.64         \\
		$\times$ &$\times$         &$\checkmark$        &15.48          &47.11         &61.28 \\
		$\checkmark$&$\checkmark$        &$\times$       &38.01          &62.03         &70.58         \\
		$\checkmark$ &$\times$         &$\checkmark$       &26.58          &57.23         &68.67 \\
		$\times$ &$\checkmark$         &$\checkmark$       &23.28          &56.39         &68.25 \\
		$\checkmark$ &$\checkmark$         &$\checkmark$       &\textbf{38.51}          &\textbf{62.33}         &\textbf{70.78} \\
		\bottomrule        
	\end{tabular}
 \caption{Top-1 accuracy (\%) of DeiT-T on ImageNet.} \label{tab:ablation-overall}
\end{table}

To demonstrate the effectiveness of the key components in AIQViT, we conduct ablation studies on the ImageNet dataset with DeiT-T. For convenience,
the architecture-informed low-rank compensation, dynamic focusing quantizer, and curriculum learning strategy are abbreviated as AILoC, DFQ, and CL, respectively. Quantitive experimental results are detailed in Table~\ref{tab:ablation-overall}. Note that, when DFQ is excluded, a uniform quantizer is employed for post-Softmax activations. The results indicate that AIQViT obtains the best results when all the variants are used. Specifically, compared with the vanilla (all variants are excluded), AILoC improves the test accuracy by 15.31\%, 10.80\%, and 7.61\% for W3/A3, W4/A4, and W6/A6 quantization, respectively, confirming the effectiveness of the low-rank compensation mechanism used in AILoC. Besides, AIQViT suffers from an 11.93\% accuracy drop when DFQ is absent, claiming the superiority of DFQ in handling low-bit cases. We also observe that the CL strategy brings more significant improvement for low-bit quantization than high-bit quantization. This can be attributed to the fact that the low-bit model benefits more from the smoother optimization objective used in CL.

To validate the superiority of architectures searched in AILoC, we conduct experiments on ImageNet with DeiT-T and DeiT-S. The results are provided in Table~\ref{tab:rank}. AIQViT with automatic $r$ consistently performs better than those with fixed $r$. This is mainly due to the differentiable architecture search, which brings more suitable architectures for network quantization. For DeiT-S, in the cases of W4/A4 and W6/A6 quantization, models with $r=20$ outperform those with $r=100$ by 1.0\% and 0.3\%, which show that directly increasing $r$ can not ensure better performances.
\begin{table}[!t]
	\centering
	
	\begin{tabular}{clccc}
		\toprule
		\multirow{2}{*}{\textbf{Model}}     & \multirow{2}{*}{\textbf{Method}} & \multicolumn{3}{c}{\textbf{Bits.(W/A)}} \\
		
		&                         & 3/3      & 4/4     & 6/6     \\
		\midrule
		\multirow{3}{*}{DeiT-T} 
		& $r$=20                    &33.78          &59.61         &68.99         \\
		& $r$=100                   &36.87          &61.18         &69.92         \\
		& Auto $r$                  &\textbf{38.51}          &\textbf{62.33}         &\textbf{70.78}         \\
		\midrule
		\multirow{3}{*}{DeiT-S} 
		& $r$=20 &50.58                   &69.73          &77.31                  \\
		& $r$=100                   &53.64          &68.73         &77.01        \\
		& Auto $r$                  &\textbf{55.36}          &\textbf{72.75}         &\textbf{78.98}         \\
		\bottomrule
	\end{tabular}
 \caption{Results of DeiT-T on ImageNet with different $r$. ``Auto $r$'' means setting $r$ by network architecture search.} \label{tab:rank}
\end{table}

\begin{figure}[!h] 
	\centering 
	\includegraphics[width=0.46\textwidth]{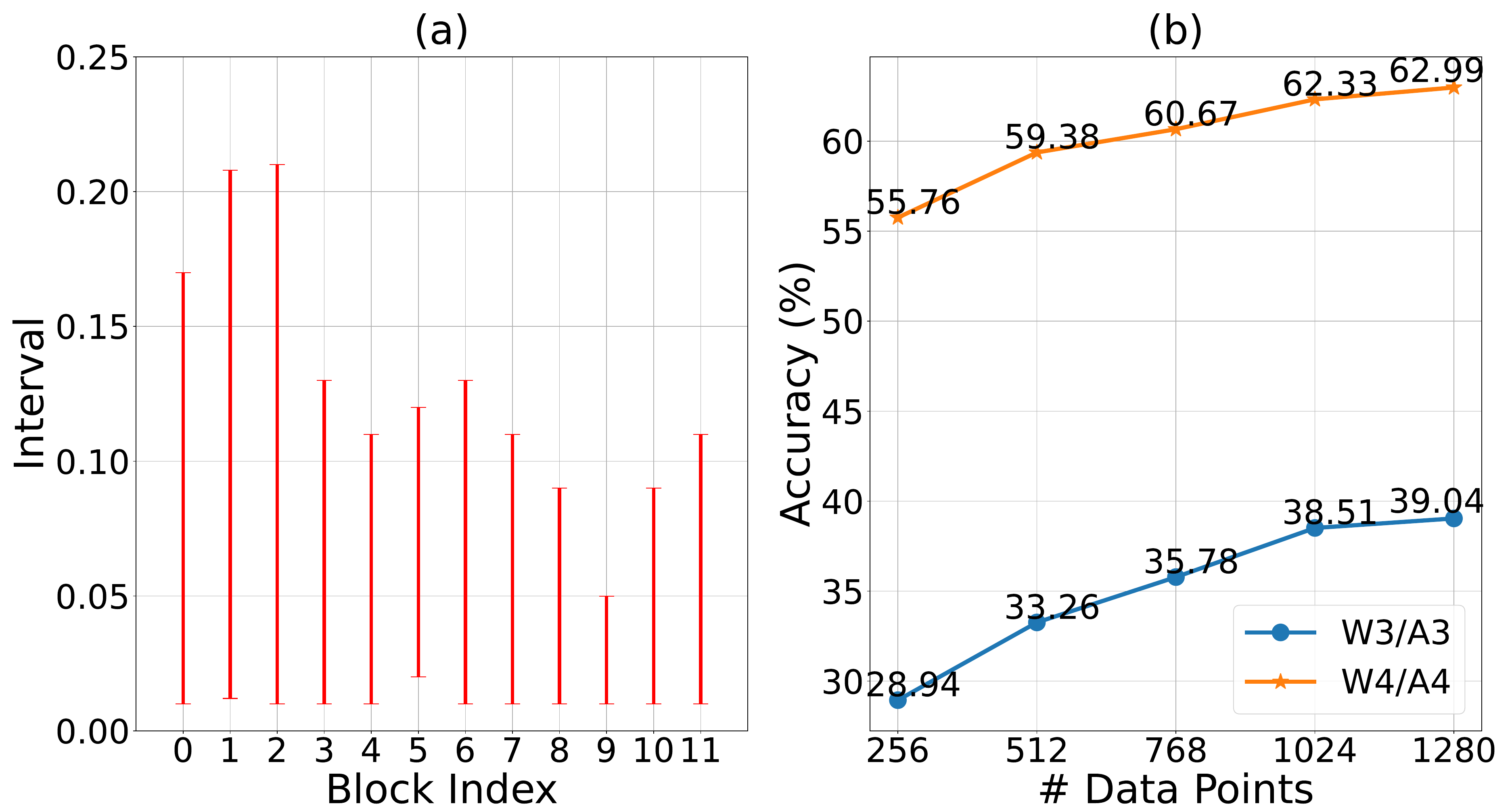} 
	\caption{ Visualization of the learned intervals and the influences of calibration data size. (a) Visualization of learned intervals for DeiT-T with W4/A4 quantization. (b) Effect of \# data points for DeiT-T quantization on ImageNet.} \label{fig:data_points}
\end{figure}

Figure~\ref{fig:data_points}(a) illustrates the quantization intervals learned by DFQ. The results reveal that different layers correspond to distinct intervals, with a notable trend that shallow layers tend to have larger intervals compared to deep layers. We also investigate the impact of calibration dataset size. As shown in Figure~\ref{fig:data_points}(b), W4/A4 quantization exhibits better robustness to the calibration size compared to W3/A3 quantization. When only 256 samples are used, both W4/A4 and W3/A3 quantization yield subpar results, which can be attributed to the overfitting led by the limited data.

\section{Conclusion}
This paper proposes AIQViT, a post-training quantization method tailored for ViTs. AIQViT employs an architecture-informed low-rank compensation mechanism, which uses the network architecture search and the CL strategy for rank calculation and stable optimization, respectively. Additionally, a simple yet effective DFQ is proposed to address the unbalanced distributions of post-Softmax activations instead of the less efficient logarithmic operations, and thus further improves the quantization efficiency. Experiments on five vision tasks demonstrate that the cumbersome ViTs can be compressed into their low-bit equivalents without compromising performances. 
In future works, we plan to develop novel PTQ methods for large models, hopefully extending these models to mobile applications.

\section*{Acknowledgments}
This work was partially supported by the National Natural Science Foundation of China (No. U20A20185, 62372491), the Guangdong Basic and Applied Basic Research Foundation (2022B1515020103, 2023B1515120087), the Shenzhen Science and Technology Program (No. RCYX20200714114641140).

\bibliography{aaai25}

\end{document}